\documentclass[10pt,twocolumn,letterpaper]{article}

\usepackage{wacv}
\usepackage{times}
\usepackage{epsfig}
\usepackage{graphicx}
\usepackage{amsmath}
\usepackage{amssymb}
\usepackage{color}
\usepackage{float}
\usepackage{subfig}
\usepackage{multirow}
\usepackage[linesnumbered,ruled]{algorithm2e}
\usepackage{comment}
\usepackage{url}

\newcommand{\vc}[1]{\bold #1}
\newcommand{\Real}{\mathbb R}
\newcommand{\Integer}{\mathbb Z}
\newcommand{\qq}[1]{\dot{{\bold #1}}}
\newcommand{\norm}[1]{\left\Vert#1\right\Vert}

\usepackage[pagebackref=true,breaklinks=true,letterpaper=true,colorlinks,bookmarks=false]{hyperref}

\wacvfinalcopy 
\def\wacvPaperID{151} 

\ifwacvfinal\fi
\begin{document}

\title{Resultant Based Incremental Recovery of Camera Pose from Pairwise Matches}

\author{Yoni Kasten \hspace{2cm} Meirav Galun  \hspace{2cm}Ronen Basri \\
Weizmann Institute of Science\\
{\tt\small  \{yoni.kasten,meirav.galun,ronen.basri\}@weizmann.ac.il}
}

\maketitle
\ifwacvfinal\thispagestyle{empty}\fi

\begin{abstract}
  Incremental (online) structure from motion pipelines seek to recover the camera matrix associated with an image $I_n$ given $n-1$ images, $I_1,...,I_{n-1}$, whose camera matrices have already been recovered. In this paper, we introduce a novel solution to the six-point online algorithm to recover the exterior parameters associated with $I_n$. Our algorithm uses just six corresponding pairs of 2D points, extracted each from $I_n$ and from \textit{any} of the preceding $n-1$ images, allowing the recovery of the full six degrees of freedom of the $n$'th camera, and unlike common methods, does not require tracking feature points in three or more images. Our novel solution is based on constructing a Dixon resultant, yielding a solution method that is both efficient and accurate compared to existing solutions. We further use Bernstein's theorem to prove a tight bound on the number of complex solutions. Our experiments demonstrate the utility of our approach.

\end{abstract}
\makeatletter
\def\blfootnote{\xdef\@thefnmark{}\@footnotetext}
\makeatother
\section{Introduction}
This paper addresses the problem of recovering camera position and orientation in a new image \(I_{n}\) given a stream of $n-1$ images $I_1,...,I_{n-1}$ whose parameters are known. Efficient and accurate solutions to this problem are important in a variety of online applications such as online robot positioning, incremental structure recovery from video images, and streaming applications.

Incremental structure from motion (SFM) pipelines \cite{snavely2008modeling,agarwal2011building,wu2011visualsfm} commonly compute the next camera pose by matching triangulated points (i.e., points whose depths have been recovered) from \(I_1,...,I_{n-1}\) to image points from \(I_n\), e.g., using PnP algorithms \cite{ameller2000camera,gao2003complete,lepetit2009epnp,camposeco2017toroidal}.  This process requires tracking feature points in three or more images. Alternative methods compute this pose by integrating information from two (or more) essential matrices (e.g., \cite{jiang2013global}). This approach utilizes an excess number of matching pairs --  at least 10 pairs are needed to recover two essential matrices. Recent work \cite{stewenius2005solutions,zheng2015structure} showed that six pairs of points suffice to determine the next camera pose. Such matching pairs can relate the new image \(I_{n}\) to \textit{any} of the previous images \(I_1,...,I_{n-1}\) (in particular each matching point may relate \(I_{n}\) to a different image). 

\begin{figure}[t]

\centering
\includegraphics[width=0.8\linewidth]{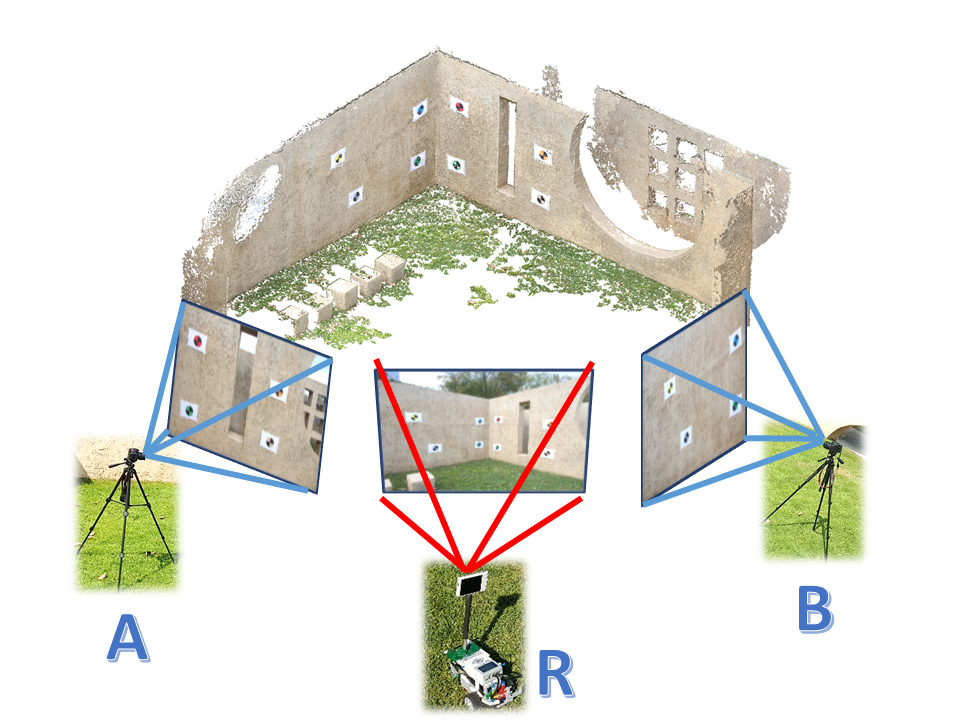} \ \ 
\caption{ \small
Images of an outdoor scene containing an \textit{internal} wall corner with eight landmarks. Stationary cameras A and B, whose parameters are known, produce non-overlapping images. A robot mounted camera (R), whose parameters are unknown, is moving in the scene. Our method is used to recover the robot position and orientation from R-A and R-B matching pairs. 
}
\label{figure:robot_images}
\end{figure}

Figure~\ref{figure:robot_images} illustrates this setup. The figure shows a scene with eight landmarks placed on two walls pictured by three cameras. Each of the (fully calibrated) cameras A and B is positioned such that it sees only four of the landmarks on one of the walls. The third, robot mounted camera, whose pose is unknown, sees all the landmarks. Our goal is to recover its pose. Note that none of the points can be triangulated because the fields of view of the two calibrated cameras are non-overlapping. Also, it is not possible to recover the essential matrices that involve the third camera since they share only four landmarks with cameras A and B, and moreover these landmarks are coplanar. Nevertheless, with six pairs of matching landmarks it is possible to recover the pose of the third camera.

Below we introduce a novel solution to the online, 6-point algorithm to recover the position and orientation of a new camera in sequential, multiview SFM. We use quaternions to obtain a succinct polynomial system in the unknown pose parameters. We compute the mixed volume of the associated Newton polytope to prove analytically that the system gives rise to 64 complex solutions and show empirically that it typically produces roughly 23 real solutions. This rigorous argument confirms previous observations based on random assignments of coefficients \cite{stewenius2005solutions}. We next symbolically derive a solution method by constructing a Dixon resultant, which we implement efficiently. Our numerical experiments indicate that our methods produces significantly more accurate solutions than efficient,  Gr\"obner base solutions \cite{stewenius2005solutions,zheng2015structure}. (\cite{byrod2009fast}'s solutions are nearly as accurate as ours, but this method is significantly slower.) We further demonstrate the utility of our approach in sequential SFM pipelines. 

\subsection{Related work}
Stewenius et al. \cite{stewenius2005solutions} introduced an approach for recovering the relative pose of ``generalized cameras",  where a generalized camera can have multiple focal centers whose relative positions are known.
Our problem is a special case of their setting, since the known \(n-1\) cameras can be considered a single generalized camera. They further derived a solution for their system of polynomial equations (which is different from our system) by constructing a Gr\"obner basis. They then used this basis to construct an action matrix whose eigen-system returns the 64 complex solutions of the problem. Later work by Larsson et al. \cite{larsson2017efficient,larsson2018beyond}, inspired by~\cite{kukelova2008automatic}, introduced an automatic generator of action matrices, which they applied to  Stewenius et al.'s formulation. 

Both solvers \cite{stewenius2005solutions,larsson2017efficient} approach the problem by formulating an expanded set of polynomials $C\vc{x}=0$, where $C$ is called an elimination template matrix and $\vc{x}$ is the vector of all monomials occurring in these polynomials.  The action matrix is the bottom-right sub-matrix of  $C$  obtained after  applying Gauss-Jordan elimination to $C$.  This elimination process may occasionally produce a numerically unstable  action matrix, resulting in numerical inaccuracies, large residuals, and subsequently large pose errors. Byrod et al. \cite{byrod2009fast} attempted to rectify these numerical instabilities by first constructing yet a larger template matrix with more polynomials and additional monomials. They replaced elimination by the more stable $QR$ decomposition with column pivoting to produce a stable action matrix. Their method however, is problem instance dependent, and so their implementation is considerably slower than previous methods -- it is in fact as slow as iterative methods such as homotopy continuation.

The solvers, mentioned above, can handle most six-point configurations, but, as noted in \cite{zheng2015structure}, they may  become degenerate when more than three points come from a single known camera. Zheng and Wu \cite{zheng2015structure} addressed this special case. Their approach is related to \cite{stewenius2005solutions} and is  subject to similar inaccuracies. Also related are generalized PnP algorithms \cite{camposeco2016minimal,sweeney2014gdls}, which align a 3D object to multiple cameras.



\label{sec:introduction}
\section{Method}
\label{section:method}
We present our method in this section. We introduce our polynomial system in Sec.~\ref{sec:definition}, convert it to a quaternion representation (Sec.~\ref{sec:quat_rep}), prove an upper bound on the number of complex solutions (Sec.~\ref{section:num_solutions}), and finally derive its Dixon resultant to develop an efficient solution method (Sec.~\ref{section::solving}).
 
\subsection{Problem definition}  \label{sec:definition}
Our input consists of point correspondences extracted from $n$ input images, $I_1,\ldots,I_n$, ($n\geq3$). We assume that all cameras are internally calibrated and, further, that the exterior parameters of the first $n-1$ cameras in a global coordinate system are known, i.e., the camera positions, $\vc{t}_i \in \Real^3$, and orientations $R_i \in SO(3)$ ($1 \le i \le n-1$), are given. Our goal is to recover the position, $\vc{t}_n$, and orientation, $R_n$, used to produce the new image, $I_n$. Below we denote our unknowns $R=R_n$ and $\vc{t}=\vc{t}_n$, omitting their subscript to simplify notation. 

To determine the 6 degrees of freedom in $R$ and $\vc{t}$ we use 6 pairs of corresponding points $\{\vc{p}_k,\vc{p}'_k\}_{k=1}^6$, where $\vc{p}_k \in I_n$, $\vc{p}'_k \in I_{i_k}$, and $1\le i_k \le n-1$. We further assume that $|\bigcup_{k=1}^6 \{i_k\}|  \ge 2$, i.e., that the corresponding points relate points in $I_n$ to at least two of the first $n-1$ images, and that together with $I_n$, these images are produced by cameras that are not all collinear. (We further discuss collinear camera settings in Sec.~\ref{sec:root}.) Each corresponding pair is then related by the appropriate essential matrix $\vc{p}_k^T E_{n,i_k} \vc{p}'_k = 0$, where we denote by $E_{n,i_k}$ the essential matrix between $I_n$ and $I_{i_k}$. To simplify our notations we next omit the subscript $k$, so we write this relation as
\begin{align}
\label{eq:basic_epipolar}
\vc{p}^T E_{n,i} \vc{p}' = 0.
\end{align}

Our approach relies on expressing the essential matrix $E_{n,i}$ in terms of the camera positions and orientations (expressed in a global coordinate frame) in $I_n$ and $I_i$. Using derivation introduced in~\cite{arie2012global}:
\begin{align}
\label{eq:mica_formulation}
E_{n,i} = R^T(T-T_i)R_i.
\end{align}
$T=[\vc{t}]_\times$ and $T_i=[\vc{t}_i]_\times$ respectively are skew-symmetric matrices representing cross products with $\vc{t}$ and $\vc{t}_i$. Plugging \eqref{eq:mica_formulation} into \eqref{eq:basic_epipolar} and collecting known quantities we obtain 
\begin{align}\label{eq:basic_equation}
\vc{p}^T R^T(\vc{s} \times \vc{t} + \vc{b})=0,
\end{align}
where $\vc{p}$, $\vc{s} = -R_i \vc{p}'$, and $\vc{b} = -T_i R_i \vc{p}'$, are known quantities, while $R$ and $\vc{t}$ are unknown. Each of the six corresponding pairs contributes one equation of the form~\eqref{eq:basic_equation}. This equation is bilinear in the unknowns $R$ and $\vc{t}$ and homogeneous in $R$ (but not in $\vc{t}$). Restricting $R$ to be rotation further provides a quadratic constraint of the form $R^T R = I$, where $I$ denotes the identity matrix.

We note the difference between~\eqref{eq:basic_equation} and the usual essential matrix relation for two images that is solved in \cite{horn1991relative,nister2004efficient,li2006five}. With two images we can set the global coordinates to coincide with those of $I_i$, obtaining $R_i=I$ and $\vc{t}_i=0$, which yields $\vc{s}=-\vc{p}'$ and $\vc{b}=0$. This makes \eqref{eq:basic_equation} homogeneous in the translation parameters, $\vc{t}$. Consequently, $\vc{t}$, and likewise $E_{n,i}$, can be recovered only up to scale, requiring a mere 5 pairs of points. The use of additional images makes our equations inhomogeneous in $\vc{t}$, allowing us, when the cameras are not all collinear, to recover all the 6 degrees of freedom in the exterior parameters corresponding to $I_n$.

\subsection{Quaternion representation}
\label{sec:quat_rep}
For our equations of the form~\eqref{eq:basic_equation}, similar to \cite{stewenius2005solutions,horn1991relative,emiris1994sparse}, we use quaternions \cite{hamilton1844ii} to eliminate the orthogonality constraints, $R^T R = I$. A quaternion is represented by a 4-vector $\qq{q}= (r; \vc{v}) \in \Real^4$, where $r \in \Real$ denotes its scalar (real) part and $\vc{v} \in \Real^3$ represents its vector (imaginary) part. We use the semicolon symbol to denote column concatenation. The space of quaternions is endowed with a product operation. The Hamilton product between two quaternions,  $\qq{q}_1= (r_1;\vc{v}_1)$ and $\qq{q}_2=(r_2;\vc{v}_2)$, is defined as
\begin{align}
\label{eq:hamilton_product}
{\qq q}_1 {\qq q}_2=(r_1 r_2 - \vc{v}_1^T  \vc{v}_2; ~r_1 \vc{v}_2 + r_2 \vc{v}_1 + \vc{v}_1 \times \vc{v}_2),
\end{align}
where the `$\times$' symbol denotes the cross product between two vectors in $\Real^3$.
The conjugate of $\qq{q}= (r; \vc{v})$ is defined as $\qq{q}^*= (r; -\vc{v})$ and its reciprocal by $\qq{q}^{-1}= \qq{q}^*/\norm{\qq{q}}^2$.  Rotations are represented by unit quaternions (so in particular $\qq{q}^{-1}=\qq{q}^*$). A rotation by an angle of $\theta$ about an axis $\vc{u} \in S^2$ ($S^2$ is the unit sphere in $\Real^3$), denoted $R=R(\theta,\vc{u})$, is represented by ${\qq q}=(\cos \frac{\theta}{2} ; \sin\frac{\theta}{2} {\vc u})$. Applying $R$ to a point $\vc{p} \in \Real^3$ is expressed via conjugation as follows,
\begin{align}
\label{eq:rotation_formula}
(0;R{\vc p})={\qq q}{\qq p}{\qq q}^*,
\end{align}
where we define ${\qq p} = (0;{\vc p})$.

Using a quaternion formulation, \eqref{eq:basic_equation} can be written as 
\begin{align}
\label{eq:basic_equation_q}
\mathrm{vec}(\qq{q}\qq{p}\qq{q}^*)^T(\vc{s} \times \vc{t}+\vc{b}) = 0,
\end{align}
where we denote by $\mathrm{vec}(.)$ the vector part of a quaternion. This formulation includes 7 unknowns, the four components of ${\qq q}$ and the three components of $\vc{t}$. The former are further constrained by $\norm{\qq{q}}^2=1$. However, since~\eqref{eq:basic_equation_q} is homogeneous in $\qq{q}$ this constraint can be omitted and replaced by fixing one of the entries of $\qq{q}$. Below we fix the real part of $\qq{q}$, which we denote by $q_1$, to 1. This restricts the angle of rotation $\theta$ by requiring $\cos(\theta/2) \ne 0$, i.e., $\theta \ne \pi$. This case can be handled separately, e.g., by rotating the global coordinate system.
In summary, our formulation has 6 unknowns, the vector part of $\qq{q}$ and the components of $\vc{t}$, and so it can be solved by providing 6 corresponding pairs of points, each supplies one polynomial equation. We note that these polynomials are cubic; they are quadratic in $\qq{q}$ and linear in $\vc{t}$. 

A further change of variables, similar to a formulation suggest by Horn \cite{horn1991relative} for essential matrices, can be applied to reduce the degree of these polynomial equations to 2. This will serve us to prove a tight bound on the number of solutions and will be useful in devising an efficient solution scheme. To define the change of variables note that the first term in \eqref{eq:basic_equation_q} represents a triple product of the form $\mathrm{vec}(\qq{q}\qq{p}\qq{q}^*)^T(\vc{s} \times \vc{t})$. Triple products are invariant to cyclic permutations, and so it can be replaced by $\vc{s}^T (\vc{t} \times \mathrm{vec}(\qq{q}\qq{p}\qq{q}^*))$. By the properties of the Hamilton product, let $\qq{t}=(0; \vc{t})$, we can write
\begin{equation}
\vc{t} \times \mathrm{vec}(\qq{q}\qq{p}\qq{q}^*) = \mathrm{vec}(\qq{t} \qq{q}\qq{p}\qq{q}^*).
\end{equation}
We now let $\qq{d}=\qq{t}\qq{q}$, then \eqref{eq:basic_equation_q} becomes
\begin{equation}\label{eq:dsystem}
\vc{s}^T\mathrm{vec}(\qq{d}\qq{p}\qq{q}^*) + \vc{b}^T \mathrm{vec}(\qq{q}\qq{p}\qq{q}^*) = 0.
\end{equation}
This polynomial is quadratic in $\qq{q}$ and bilinear in $\qq{q}$ and $\qq{d}$. Note that $\qq{d}$ has 4 entries, increasing the number of variables to 7. The existence of an additional constraint is therefore implied. By the definition of $\qq{d}$, we obtain that $\qq{t}=\qq{d} \qq{q}^{-1}$.  The following relation ensures that the real part of ${\qq t}$ will be identically 0,
\begin{equation}\label{eq:dsystem_constraint}
\qq{d}^T \qq{q} = 0.
\end{equation}
We obtain in total a system of 7 polynomial equations  in 7 unknowns; $6$ corresponding pairs yield $6$ equations of type \eqref{eq:dsystem}, along with the additional constraint \eqref{eq:dsystem_constraint}.
 
\subsection{Number of solutions}
\label{section:num_solutions}

Next, we wish to determine the number of solutions to our polynomial system of equations. Previous work \cite{stewenius2005solutions,larsson2017efficient} used random coefficient assignments to argue that the problem gives rise to 64 complex solutions. Here we use Bernstein's bound \cite{bernshtein1975number} to prove this rigorously, confirming these early observations.

 One way to obtain a bound on the number of complex solutions is by using Bezout's Theorem \cite{cox2006using}. According to this theorem, a generic system of $m$ polynomials in $m$ variables of  degree $d$ should have up to $d^m$ common complex roots. Applying this to \eqref{eq:basic_equation_q}, which consists of 6 cubic polynomials yields a bound of $3^6=729$ solutions. A tighter bound is obtained using the formulation of \eqref{eq:dsystem} and \eqref{eq:dsystem_constraint}. This system consists of 7 quadratic polynomials, yielding a bound of $2^7=128$ complex solutions.

Bezout's theorem allows the polynomials to include all terms up to degree $d$ with independent coefficients. In fact, a tighter bound can be derived by noting that our polynomials are sparse, in the sense that they involve only a subset of the terms. The tighter bound is obtained by applying the Bernstein-Khovanskii-Kushnirenko (BKK) Theorem \cite{bernshtein1975number}, which relies on a remarkable connection between polynomials and convex polytopes. A polynomial $f$ in $m$ variables, $\vc{x}=(x_1,\ldots,x_m)$, is a finite sum of terms of the form $c_{\vc{\alpha}} \vc{x}^{\vc{\alpha}}$, with $c_{\vc{\alpha}} \ne 0$, $\vc{\alpha}=(\alpha_1,\ldots,\alpha_m)$, and we use the multi-index notation $\vc{x}^{\vc{\alpha}}=\Pi_{i=1}^m x_i^{\alpha_i}$. The collection of vectors of powers for $f$, $\{\vc{\alpha}\} \subset {\Integer}_+^m$, record its sparsity pattern; this point set is determined only by terms with non zero $c_{\vc{\alpha}}$. The convex hull of this set of points is called the Newton Polytope of $f$, denoted $NP(f)\subset {\Real}^m$.
Consider now a system of $m$ polynomials $f_1=...=f_m=0$ in $m$ variables and their Newton Polytopes, $P_1,...,P_m$ ($P_i=NP(f_i)$). The BKK theorem uses the mixed volume of these polytopes, $MV(P_1,\ldots,P_m)$, to derive a bound on the number of non-zero complex solutions. The mixed volume is defined by the following formula
\begin{equation}\label{eq:mixed_volume} \vspace*{-0.1cm}
MV(P_1,...,P_m)=\sum_{k=1}^m(-1)^{m-k} \hspace*{-0.35cm}
\sum_{\substack{N \subseteq \{1\ldots m\}\\ |N|=k}} \hspace*{-0.3cm}
\mathrm{Volume}(\sum_{i\in N}P_i)
\end{equation}
where we use $\sum_{i \in N} P_i$ to denote the Minkowski sum of polytopes. In the special case in which all the polynomials share the same sparsity pattern, and hence the same Newton Polytope, i.e., $P_i=P$ for all $i$, then the mixed volume is given by $MV(P_1,\ldots,P_m)=m!\,\mathrm{Volume}(P)$. Indeed, \eqref{eq:basic_equation_q} defines such a polynomial system. Computing the associated mixed volume (using the PHCpack solver~\cite{verschelde1999algorithm}) yields a bound of 160 solutions. Calculation of the mixed volume \eqref{eq:mixed_volume} for our quadratic formulation \eqref{eq:dsystem}-\eqref{eq:dsystem_constraint} yields a tighter bound consisting of 64 complex solutions. (We note that \cite{stewenius2005solutions}'s formulation -- Eq. (11) in their paper -- yields a non tight Bernstein's bound of 80.)

\begin{figure}[tb]
\centering
\includegraphics[width=0.4\linewidth]{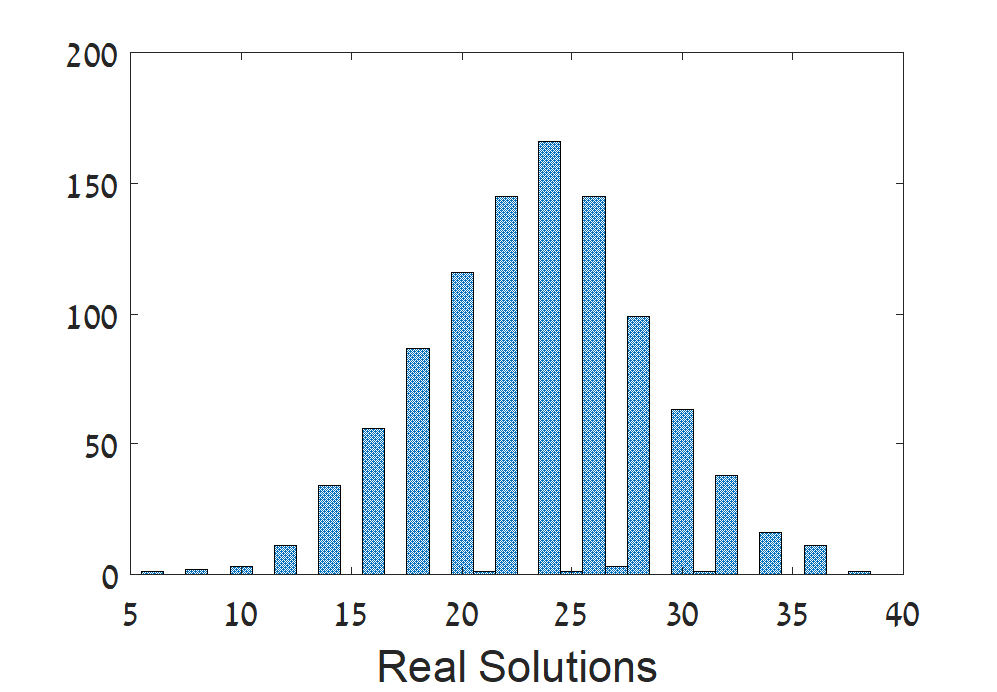}~~~~~
\subfloat{\includegraphics[width=0.4\linewidth]{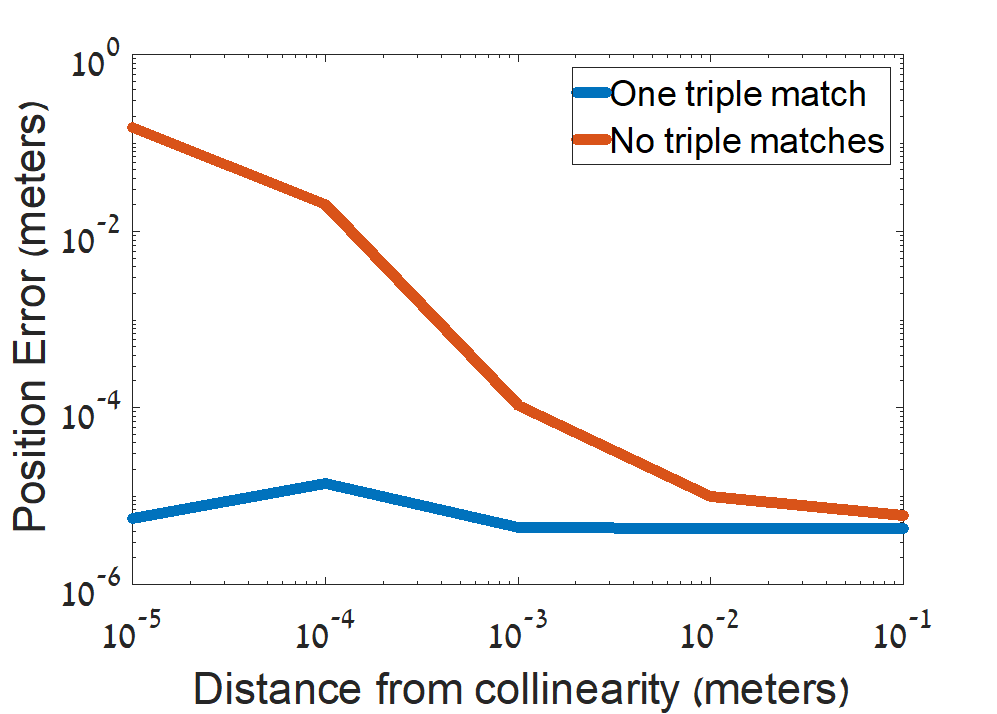}}
\caption{ \small
Left: A histogram plot showing the number of real solutions (of the 64 complex solutions). The plot represents the results of 1000 experiments in which camera settings and SIFT matches are randomly selected from the Herz-Jesus-P8 dataset. Right: Sensitivity analysis for three near collinear cameras. For a baseline of six meters between two cameras, the red curve shows the position error computed for the middle camera as a function of its deviation from the baseline. The blue line shows the position error when a single triple match is used to determine the unknown scale.}
\label{figure:hist_num_real}
\label{figure:collinear_degeneracy}
\end{figure}

We verified numerically that indeed all 64 complex solutions are attained. We used simulations to produce random camera matrices, as well as point matches, and used our method below
to numerically solve the obtained polynomial systems \eqref{eq:dsystem}-\eqref{eq:dsystem_constraint}. We obtain exactly 64 solutions in nearly all cases (due to numerical issues 0.1\% of our trials produced 62-63 solutions). (We further verified this with homotopy continuation.) 
Naturally, not all of these solutions are real. The  histogram plot in Figure~\ref{figure:hist_num_real} (left)
shows the distribution of real solutions obtained over 1000 experiments. Our experiments yielded on average $23.4 \pm 5.05$ real solutions. We are unfortunately unaware of a theoretical method to bound the number of real solutions below 64. 


\subsection{Solving the polynomial equations}
\label{section::solving}


Our next goal is to construct an efficient solution scheme to find all the real solutions of our polynomial equations, \eqref{eq:dsystem}-\eqref{eq:dsystem_constraint}. A classical method for solving such systems is by using {\em resultants}. Such solutions are typically an order of magnitude more efficient than iterative methods such as homotopy continuation. Given a polynomial system of equations, its resultant is a polynomial in the coefficients of the system that vanishes if the equations share a common root. Resultants can further be used as an effective method for variable elimination. Treating one variable as a parameter (called a {\em hidden} variable), the resultant defines a polynomial equation in that variable, eliminating simultaneously all the rest of the variables. Various approaches can then be used to solve for the hidden variable and to extend the solution to all the rest of the variables.

The algebraic geometry literature offers several ways to construct resultants. Some of these approaches, unfortunately, may produce very large resultants that are difficult to work with. In this work we chose to  use the Bezout-Cayley-Dixon (BCD) method~\cite{dixon1909eliminant}, which allows us to obtain relatively compact expressions. We then use the obtained resultant       to solve our equations by casting it as a generalized eigensystem problem in the form suggested in~\cite{manocha1994solving}.

\noindent \textbf{The BCD method}. \
Given a polynomial system with $m+1$ equations in $m+1$ variables, $x_1,...,x_{m+1}$, and assume without loss of generality that we choose to hide $x_{m+1}$, then we express the polynomials in terms of the rest of the variables $\vc{x}=(x_1,...,x_m)^T$as
\begin{equation}  \label{eq:polynomials}
f_1(\vc{x})=...=f_{m+1}(\vc{x})=0.
\end{equation}
To construct the resultant we introduce new variables $\vc{y}=(y_1,\ldots,y_m)^T$ and construct the following $(m+1)\times(m+1)$ matrix, $D({\vc x},{\vc y})=$
\begin{equation}\label{eq:dixon_matrix}
\left(\begin{smallmatrix}
f_1(x_1,x_2,...,x_m)& f_2(x_1,x_2,...,x_m) & ... & f_{m+1}(x_1,x_2,\ldots,x_m)\\
f_1(y_1,x_2,...,x_m)& f_2(y_1,x_2,...,x_m) & ... & f_{m+1}(y_1,x_2,\ldots,x_m)\\
f_1(y_1,y_2,...,x_m)& f_2(y_1,y_2,...,x_m) & ... & f_{m+1}(y_1,y_2,\ldots,x_m)\\
\vdots& \vdots & \vdots \\
f_1(y_1,y_2,...,y_m)& f_2(y_1,y_2,...,y_m) & ... & f_{m+1}(y_1,y_2,\ldots,y_m)
\end{smallmatrix} \right)
\end{equation}
so that at each row $i$, one more variable $y_i$ replaces a respective variable $x_i$. The Dixon polynomial is defined as 
\begin{equation}
\delta({\vc x},{\vc y}) = \frac{\det(D({\vc x},{\vc y}))}{(x_1-y_1)\cdot \cdot \cdot (x_m-y_m)}.
\end{equation}
$\delta(\vc{x},\vc{y})$ is indeed a polynomial; it can be verified that $\text{det}(D)$ is divisible by $(x_1-y_1)\cdot\cdot\cdot(x_m-y_m)$ by the following argument. If we subtract row $i+1$ in $D$ from row $i$ ($1 \le i \le m$; such an operation does not change the determinant) we obtain expressions of the form  $f_i(\ldots,x_i,\ldots)-f_i(\ldots,y_i,\ldots)$. Such an expression vanishes at $y_i=x_i$, and so it contains a multiple of $(x_i-y_i)$. Overall, $\delta(\vc{x},\vc{y})$ is a polynomial of degree $((i\times d_i)-1)$ in $x_i$ and $((m+1-i)\times d_i-1)$ in $y_i$ for $1\leq i \leq m$, where $d_i$ is the maximal degree of $x_i$ in $f_1,\ldots,f_{m+1}$. 

The Dixon polynomial, $\delta({\vc x},{\vc y})$, vanishes for any common root ${\vc x}$ of~\eqref{eq:polynomials}, regardless of $\vc{y}$. This is clear, because the first row of $D$, which is independent of ${\vc y}$, vanishes with any common root $\vc{x}$. If we now express $\delta({\vc x},{\vc y})$ as a sum of monomials in $\vc{y}$, i.e.,
\begin{equation}
\delta({\vc x},{\vc y}) = \sum_\alpha \big(\sum_\beta \tilde M_{\alpha,\beta}{\vc x}^{\beta}\big){\vc y}^\alpha
\end{equation}
then clearly all the coefficients of ${\vc y}^{\alpha}$, $\sum_\beta \tilde M_{\alpha,\beta}{\vc x}^{\beta}$, must vanish identically. This yields the following linear system 
\begin{equation}  \label{eq:Dixon_matrix}
\tilde M \tilde {\vc v}=0,
\end{equation} 
where $\tilde M$, called the Dixon matrix, is the matrix of the coefficients $\tilde M_{\alpha,\beta}$, and $\tilde {\vc v} = (\ldots,\vc{x}^\beta,\ldots)^T$.


%

In principle, if $\tilde M$ was square, we could solve for~\eqref{eq:polynomials} by finding values of the hidden variable that make ${\tilde M}$ singular (recall that its entries are polynomials in the hidden variable $x_{m+1}$),
and then computing the null space of $\tilde M$. In many cases, however, $\tilde M$ is rectangular and may be identically rank deficient.
Kapur et al.~\cite{kapur1994algebraic} showed that under certain conditions requiring the determinant of any maximal rank submatrix of $\tilde M$ to vanish provides a necessary condition for the existence of a common root. This way, it is possible to obtain, after elimination of rows and columns, an informative square matrix $M$. 

\noindent \textbf{Symbolic construction of the Dixon matrix.}\label{sec:symbolic} \ Our polynomials \eqref{eq:dsystem}-\eqref{eq:dsystem_constraint} consist of $7$ equations in $7$ unknowns. We chose to hide $q_2$ (recall that we set $q_1 = 1$), yielding ${\vc x}=(d_1,d_2,d_3,d_4,q_3,q_4)$,
and hence a matrix $D$ of size $7 \times 7$.  Numeric construction of the Dixon matrix can be obtained by plugging values for the 6 corresponding points into \eqref{eq:dsystem} and then applying the Maple package of \cite{minimair15} to $D$. Using the method above we obtain a Dixon matrix $M$ of size $27 \times 27$ whose entries are polynomials in $q_2$, with maximal degree $8$, and a vector ${\vc v}$ of size $27$, composed of all powers of $q_3q_4$ of degree up to 6, except $q_3^6$ (due to the order of elimination). 
Interestingly in this process all terms that involve $\vc{d}$ are eliminated, and so our resultant only depends on the rotation parameters. This will be useful when the cameras are all collinear, as we further elaborate in Sec.~\ref{sec:root}.

To speed up computations we aim to generate a symbolic expression of the Dixon matrix $M$. Direct application of this construction unfortunately yielded polynomials that were too long to store in the computer memory. (Already the determinant of a $5 \times 5$ submatrix produced an expression of size 176Mb, and its calculation required even more memory.) 
To overcome this, we recursively applied the Laplace expansion, constructing polynomials from determinants of all submatrices of $D$ of size $3 \times 3$. We then simplified these polynomials and introduced additional variables for the obtained coefficients of terms in $\vc{v}$ and the hidden variable $q_2$. We later used the obtained expressions to construct polynomials for determinants of all submatrices of size $4 \times 4$. We repeated this process until we obtained a single polynomial expression for the determinant of the full $7 \times 7$ matrix $D(\vc{x},\vc{y})$. Next, we reorganized our terms in the form of~\eqref{eq:Dixon_matrix}, obtaining a matrix $\tilde M$ of size $27 \times 184$. Finally, after removing all identically zero columns, we obtained a $27 \times 27$ Dixon matrix $M$ of the form
\begin{align}  \label{eq:Dixon_solution}
M(q_{2}) = M_0 + q_2 M_1 + q_2^2 M_2 + \dots + q_2^8 M_8,
\end{align}
where the entries of $M_i$ ($0 \le i \le 8$) are functions of known quantities, $\vc{p}$, $\vc{s}$, and $\vc{b}$, \eqref{eq:basic_equation}, which are determined by each of the 6 corresponding pairs of points given as input. Following~\eqref{eq:Dixon_matrix}, this matrix satisfies
\begin{align}
M \vc{v} = 0.
\end{align}

\noindent \textbf{Finding the roots.} \label{sec:root} $\ $To find the common roots of~\eqref{eq:polynomials} we need to find assignments of the hidden variable, $q_2$, that make $M$ singular. This can be achieved for example by calculating the roots of $\det(M)$ as a univariate polynomial of degree 64. A more stable approach is to apply an eigen-decomposition technique, following~\cite{manocha1994solving}. Given~\eqref{eq:Dixon_solution},
it can be readily verified that $q_2$ is an eigenvalue of the following generalized eigen-system
\begin{equation}  \label{eq:gen_eigen}
 C_2 {\bar {\vc v}} = q_2 C_1 {\bar {\vc v}},
\end{equation} 
with
\begin{equation}
C_1=
\left(\begin{smallmatrix}
I & 0 &\dots & 0 \\
0 & I &\dots & 0 \\
 \vdots & \vdots & \vdots &\vdots &\\ 
0&0&\dots &M_{8}  
\end{smallmatrix}\right),~ 
C_2=
\left(\begin{smallmatrix}
0& I & 0 &\dots & 0 \\
0& 0 & I &\dots & 0 \\
 \vdots & \vdots & \vdots &\vdots & \vdots\\ 
 0& 0 & 0 &\dots & I \\
 -M_0&-M_1&-M_2&\dots &-M_{7}  
\end{smallmatrix}\right)
\notag
\end{equation}
and the corresponding eigenvector ${\bar {\vc v}}$ is given by
\begin{equation}
{\bar {\vc v}}=({\vc v}; q_2 {\vc v}; q_2^2 {\vc v}; \ldots;  q_2^7 {\vc v}).
\end{equation}
The matrices $C_1$ and $C_2$ are of size $(27\times 8) \times (27\times 8)=216 \times 216$. Using our symbolic calculation we verified that $\det(M)=\det(C_2-q_2C_1)$ is a polynomial of degree 64 and that $C_1$ is singular. Consequently, of the $216$ eigenvalues of this generalized eigen-system all but $64$ of them diverge.
This is compatible with the BKK bound derived in Sec.~\ref{section:num_solutions}. Note that, since we have set $q_1=1$, a smaller number of solutions may be obtained if rotation of $\pi$  is a solution. The remaining solutions can be found by repeating the process after rotating the global coordinate system. 

Of the 64 eigenvalues we ignore complex ones. We then assign each of the remaining eigenvalues to $q_2$ and extract $q_3$ and $q_4$ from the second and third entries of the respective eigenvector ${\bar {\vc v}}$. Together we obtain the quaternion ${\qq q} =(1,q_2,q_3,q_4)$, which we then normalize to obtain a rotation quaternion. Note that there is no sign ambiguity since two unit quaternions $\qq{q}$ and $-\qq{q}$ represent the same rotation. 

Once we recover the rotation parameters we use~\eqref{eq:basic_equation_q}, which is linear in $\vc{t}$, to solve directly for $\vc{t}$. In a general camera setup this linear system is of rank 3, allowing us to solve for all the three degrees of freedom in $\vc{t}$. If this linear system is degenerate, however, it indicates that the cameras are all (near) collinear, in which case we only recover $\vc{t}$ in a line in 3D (i,e., up to scale if we set the origin of the global coordinate system at any point along the line). 

We can resolve this ambiguity in the translation by using one matching triplet, if available. Let \(\vc{p}_i \in I_i\), \(\vc{p}_j \in I_j\) \((1 \le i<j <n)\), and \(\vc{p} \in I_n\), and let \(\vc{P}\) denote the 3D point obtained by triangulating \(\vc{p}_i\) and \(\vc{p}_j\). Suppose that \(\vc{t}\) is recovered in a line, i.e., \(\vc{t}=\vc{t}'+\alpha\vc{t}''\), where both \(\vc{t}', \vc{t}'' \in \Real^3\) are known and \(\alpha\) is an unknown scalar. Then,
\begin{equation}
\vc{p}\propto R^T (\vc{P}-\vc{t})=R^T(\vc{P}-\vc{t}'-\alpha\vc{t}'').
\end{equation} 
We can determine \(\alpha\) by solving the linear system
\begin{equation}  \label{eq:collinear}
\vc{p} \times R^T(\vc{P}-\vc{t}'-\alpha\vc{t}'')=0.
\end{equation}

\begin{algorithm}[t]

    \SetKwInOut{Input}{Input}
    \SetKwInOut{Output}{Output}

    \Input{$R_1,...,R_{n-1}, {\bf t}_1,...,{\bf t}_{n-1} ,\{(\vc{p_k},\vc{p'_k})\}_{k=1}^6   $\\$s.t \ \ \vc{p}_k \in I_n, \vc{p}_k' \in I_1 \cup ... \cup I_{n-1}    $}
    \Output{ $\{U_i\}_,\{\tau_i \}$, a solution set of orientations and locations}    
        
        $M_0,\dots,M_8\leftarrow$ \it{Use our symbolic derivations to construct resultant matrices~\eqref{eq:Dixon_solution}}
        \\
        $C_1,C_2\leftarrow$ \it{Build generalized eigen-system~\eqref{eq:gen_eigen}} 
        \\ 
$\{(q_{2},q_3,q_4)\}_{\{1...64\}}\leftarrow$ Solve eigen-system~\eqref{eq:gen_eigen} \\
$\{\qq{q_i}\}\leftarrow$ \it{Normalize and keep real solutions}\\
$ \{U_i\}\leftarrow$ \it{extract orientation matrices}\\
$\{\tau_i\}\leftarrow$ \it{Solve for positions linearly using~\eqref{eq:basic_equation_q}}\\

return $\{U_i\},\{\tau_i\}$
\caption{Recover camera position and orientation for an image $I_n$ given the locations and orientations of $I_1,...,I_{n-1}$}
\label{Alg:alg}
\end{algorithm}
Note finally that with one triple match we need only 4 more pairs of points to solve for all six degrees of freedom in the exterior parameters. (See also~\cite{josephson2007image} for solutions that combine pairs and triplets.) The triple match $\vc{p},\vc{p}_i,\vc{p}_j$  provides two polynomial equations to our polynomial system~\eqref{eq:basic_equation} (for the pairs $\vc{p},\vc{p}_i$ and $\vc{p},\vc{p}_j$), allowing with the additional 4 pairs to solve for 5 of the DOFs in the camera matrix. The same triple match can be used further to solve for the missing scale using~\eqref{eq:collinear}.

Figure~\ref{figure:collinear_degeneracy} (right) shows the error in recovering camera position for near collinear camera setups. Indeed, in this case, with just pairwise correspondences  the position of the recovered camera is determined only up to the scalar $\alpha$. However, using a single matching triplet this scalar is recovered accurately even for exactly collinear cameras. Our method is  summarized in Alg.\ref{Alg:alg}.
 \begin{table*}[t]
\caption{\small 
Runtime, orientation recovery error and residual obtained with our method compared to existing methods. }
   \centering \scriptsize
   \begin{tabular}{|l|l||c|c|c|c||c|c|}
     \hline
   \multicolumn{2}{|c||}{~}& \multicolumn{4}{|c||}{\textbf{General Case}}& \multicolumn{2}{|c|}{\textbf{4+2}}\\
   \cline{3-8}
   \multicolumn{2}{|l||}{} & \textbf{Ours} &
   \textbf{Byrod et al.} & \textbf{Larsson et al.} & \textbf{Stewenius et al.}& \textbf{Ours}& \textbf{Zheng et al.} \\
  \hline
   \multicolumn{2}{|l||}{ Runtime (ms)}   & 21.4 & 260.1 & \textbf{1.7} & 2.3&20.3&\textbf{1.4} \\ \cline{1-8}
   \multirow{2}{7em}{Rotation Error (degrees)}
  & mean & \textbf{6.3096e-07} & 9.7899e-04 &0.2457& 1.8051 &\textbf{0.0041}&4.2196\\ \cline{2-8}
   &median &\textbf{7.6592e-09}  &5.9371e-08 &3.3821e-05&  0.1330&\textbf{7.1092e-08}&3.2804e-04  \\ \cline{1-8} 
   
   \multirow{2}{5em}{Stewenius' residual}
  & mean & \textbf{4.3453e-10} &  7.1570e-08 &9.7387e-04& 0.0045& \textbf{5.3624e-07}& 9.9476e+09\\ \cline{2-8}
   &median &\textbf{3.5269e-12}  &4.2803e-11 &1.8555e-08&5.5276e-05 &\textbf{6.5168e-11} &1.3399e-07\\ \cline{1-8}
\end{tabular}
\label{tab:runtimeTable}

\end{table*}
\begin{table*}[tb]
\caption{\small
 Accuracy of pose recovery from landmarks. Results shown are averages over 22 different camera locations.}
\begin{center}\vspace{-2mm}
   \centering
   {\scriptsize
   \begin{tabular}{|l|l|l||c|c|c|c||c|c|}
   \hline
   \multicolumn{3}{|l||}{}& \multicolumn{4}{|c||}{3+3}& \multicolumn{2}{|c|}{4+2}  \\\cline{4-9}
   
   \multicolumn{3}{|l||}{} & \textbf{Ours} &
   \textbf{Byrod et al.} & \textbf{Larsson et al.} & \textbf{Stewenius et al.}&\textbf{Ours}&\textbf{Zheng et al.} \\
   \hline
   \multirow{2}{5em}{Position error} & \multirow{2}{4.2em}{Meters} & mean & \textbf{0.0278} & 0.0323 & 0.5418&0.9837 &\textbf{0.0148}&0.1164\\ \cline{3-9}
   && median &\textbf{0.0286} &\textbf{0.0286} & 0.0388& 0.1022&\textbf{0.0132}&0.0135  \\
   \hline
   \multirow{4}{4.3em}{Orientation error} & \multirow{2}{3em}{Frobenius}
   & mean & \textbf{0.0063} & \textbf{0.0063} &0.1657& 0.2213 &\textbf{0.0050}&0.1097\\ \cline{3-9}
   && median &\textbf{0.0064}  &\textbf{0.0064} &0.0074& 0.0223 &\textbf{0.0048}&0.0050\\ \cline{2-9}
   & Degrees & mean & \textbf{0.2538} & \textbf{0.2538} & 6.8315& 9.0557 & \textbf{0.2020}&5.165\\ \cline{3-9}
   && median &\textbf{0.2599} &\textbf{0.2599} & 0.3002& 0.9037&\textbf{0.1938}&0.2010\\
   \hline
 \end{tabular}
 }
 \end{center}
 
\label{tab:landmarks_33}  
\vspace{-7mm}
\end{table*}

\noindent \textbf{More than 3 correspondences from one camera.} $\ $
The $27 \times 27$ Dixon matrix $M $  constructed in~\eqref{eq:Dixon_solution} becomes singular when four of the six matching pairs come from a single image $I_i$ ($1 \le i \le n-1$) (and the other two from either one or two images), in which case its rank for a general assignment of $q_2$ is 23. Following Kapur et al.'s main theorem~\cite{kapur1994algebraic}, we use the sub-matrix $\overline{M}=M_{1:23,1:23}$ consisting of the first $23$ rows and columns of $M$ which is generally full rank. The solutions for $q_2$ can be  found by applying the respective generalized eigen-system for $\overline{M}$. 

To solve for the remaining variables $q_3$ and $q_4$, we note first that, unlike in the previous case, the obtained eigenvector $\overline{\vc{ {v}}}$, which  satisfies $\overline{M}\overline{\vc{ {v}}}=0$,   cannot be used directly toward this goal. Instead, we plug in each of the solutions to \(q_{2}\) into \(M(q_2)\), obtaining a matrix of rank 22, and reorder its columns so that the columns that correspond to the monomials $[1,q_4,q_4^2,q_4^3,q_4^4,q_4^5]^T$ of $\vc{v}$ are placed on the right of $M$. We next apply the LU\ decomposition to \(M\), so that \(M(q_2)\vc{v}=LU\vc{v}=0\) which implies that \(U\vc{v}=0\). All but the 5 right most entries of the $22^{th}$ row of \(U\) are zeros. Multiplying this row by $\vc{v }$ (permuted accordingly) results in a 5-degree polynomial in the single variable $q_4$. For each of the 5 obtained solutions for $q_4$ we once more reorder  the columns of $M(q_2)$ this time placing the columns corresponding to the monomials $[1,q_4,q_4^2,q_4^3,q_4^4,q_3]^T$ of $\vc{v}$ on the right. We next use the LU decomposition to obtain a linear equation in $q_3$. We finally plug in the 5 solutions for \(q_3\) and \(q_4\) into our original system of equations and discard all but the solution of minimum residual. (In practice for stability, we keep the solution that minimizes the Sampson error for the 6 corresponding points.)
This case was solved in \cite{zheng2015structure} and demonstrated empirically   40  solutions. We note, however, that   \cite{zheng2015structure}'s formulation (Eqs. (3-5) in their paper) yields a non tight BKK bound, which is 56. In our formulation, we achieve a tight BKK bound, which is 40,   by placing the origin at the center of the camera which has 4 correspondences with the new camera.     

We note finally that, as is mentioned in \cite{zheng2015structure}, when 5 correspondences come from the same camera we can first recover the essential matrix relating this camera to the new camera. This yields 20 solutions, which we can use to recover 5 of the 6 degrees of freedom in the pose of new camera. We then use the remaining matching pair to recover the remaining parameter (scale of translation).

\section{Experiments}

\subsection{Runtime and accuracy}

We tested our solution in simulations and on real data. To assess its accuracy we used the  Herz-Jesus-P8 dataset \cite{strecha2008benchmarking} to generate 1000 different configurations of six pairs of points (true matches) that include 1-3 matches from up to 6 cameras. We further generated 1000 6-pair configurations that involve 4 matches from one known camera. We compare our results for the former collection with results obtained with \cite{stewenius2005solutions,byrod2009fast,larsson2017efficient} and on the latter with \cite{zheng2015structure}. (Recall that the solutions in \cite{stewenius2005solutions,byrod2009fast,larsson2017efficient} are degenerate when 4 matches come from one camera, while \cite{zheng2015structure} only addressed that special case.)

\begin{figure}[t!]
\centering
\subfloat{
\includegraphics[width=0.4\linewidth]{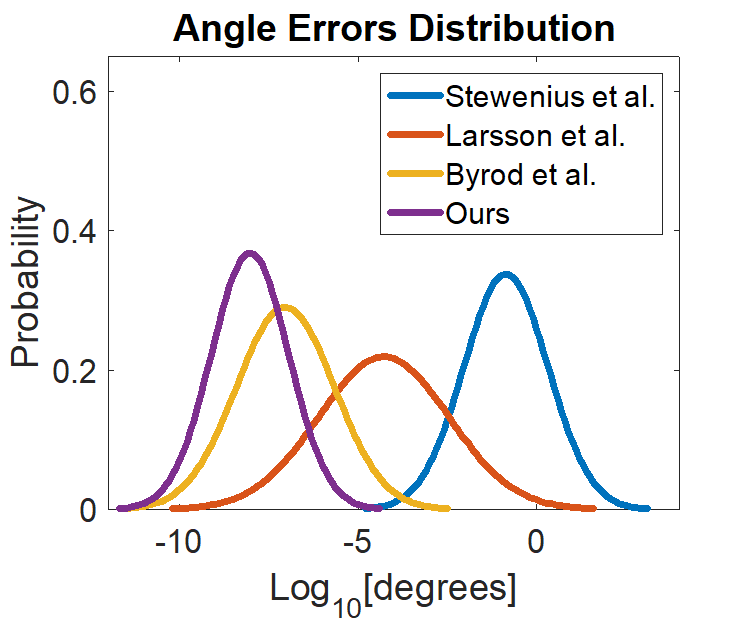}\label{fig:noise_freea}}
\subfloat{
\includegraphics[width=0.4\linewidth]{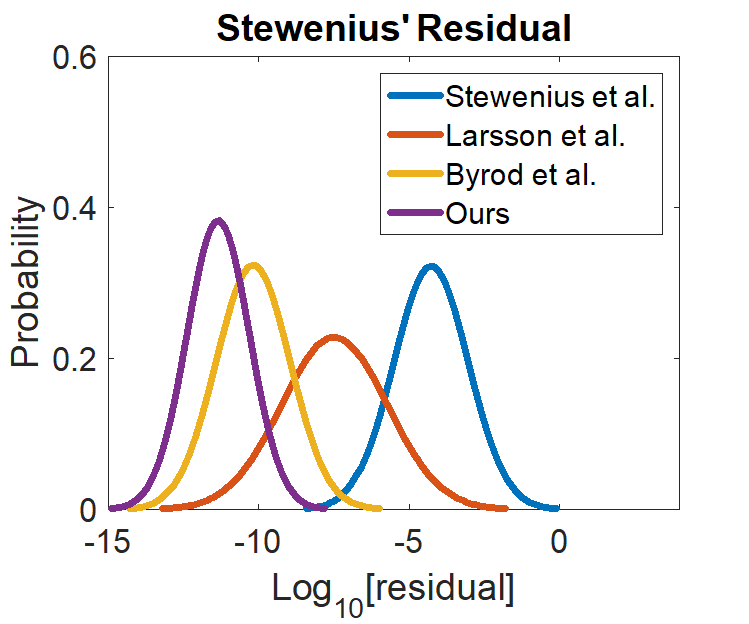}\label{fig:noise_freeb}}\\
\subfloat{
\includegraphics[width=0.4\linewidth]{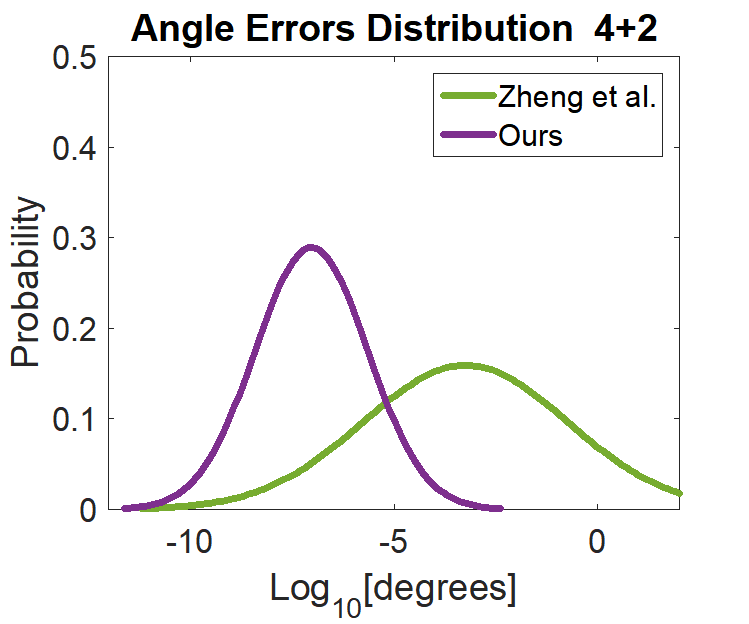}\label{fig:noise_freec}}
\subfloat{
\includegraphics[width=0.4\linewidth]{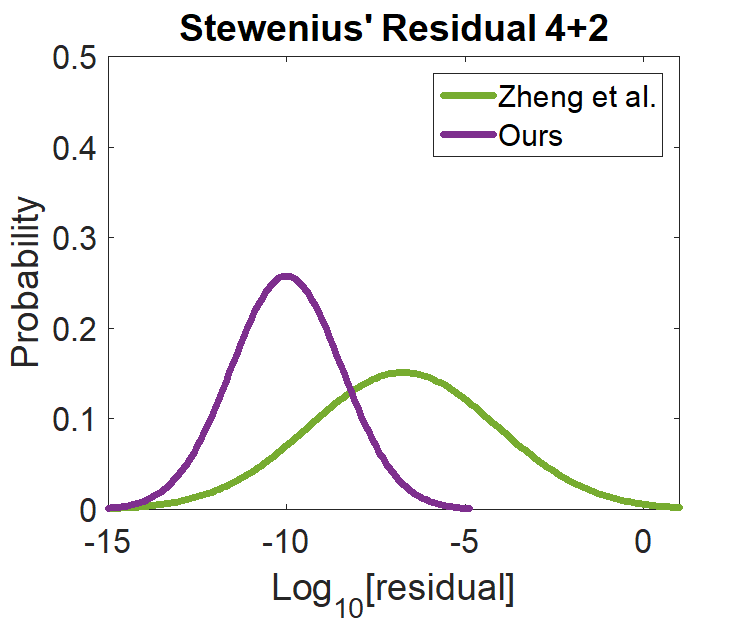}\label{fig:noise_freed}}
\caption{ \small
Distributions of orientation error and residuals produced (log10 scale) with our method compared to existing methods in 1000 trials.  }
\label{fig:noise_free}
\vspace{-10pt}
\end{figure}

The results are presented in Figure~\ref{fig:noise_free} and are summarized in Table~\ref{tab:runtimeTable}. In each case we show the recovery error of camera orientation along with the residual error obtained when each solution (including ours) is plugged into the system of 30 equations of Stewenius et al.~\cite{stewenius2005solutions} in $q_2$, $q_3$ and $q_4$. (We do not show the errors in the location parameters since in all methods those are solved linearly once the orientation parameters are recovered.) Our method achieves highly accurate results outperforming these existing methods.

Our implementation utilizes C code to construct our resultant. The code obtains six candidate matches and their respective camera matrices, as described in Sec. \ref{sec:definition}, and uses our symbolic expressions (Sec.~\ref{sec:symbolic}) to produce the nine $27 \times 27$ matrices, $M_0,...,M_8$~\eqref{eq:Dixon_solution}. We compiled the code as a Matlab Mex library and used MATLAB to solve the generalized eigen-system described in Sec.~\ref{sec:root}. Our solution runs in 20.3-21.4ms on a PC with i7-6700 3.4GHz CPU, which is compatible with real-time applications. Extracting Dixon matrix takes only 6ms of the 20ms,  much faster than the general implementation \cite{minimair15} which takes about 17 seconds. Solving the generalized eigen-system of the sparse matrices $C_1$ and $C_2$  of size $216 \times 216 ~\eqref{eq:gen_eigen}$ takes 14ms of the 20ms.    Our method is not as fast as the less accurate methods in \cite{stewenius2005solutions,larsson2017efficient,zheng2015structure}, but is significantly faster (and still more accurate) than \cite{byrod2009fast} which is incompatible with real-time applications, see Table~\ref{tab:runtimeTable}. For comparison, solving our equations with homotopy continuation takes 271ms using the multicore version of the PHCpack solver \cite{verschelde1999algorithm}.
All these methods were run on the same PC.

\begin{table*}[t]
\caption{\small Accuracy of exterior parameter estimation using our sequential SFM pipeline compared to visualSFM. The table shows the mean and median over the cameras' parameters error, where the error for each camera is averaged over 10 different runs.} \vspace{-11pt}
  \begin{center}{\scriptsize \centering
  \begin{tabular}{|l|l|l||c|c||c|c||c|c|}
  \hline
  \multicolumn{3}{|l||}{~} & \multicolumn{2}{|c||}{\textbf{Fountain-P11}}
  & \multicolumn{2}{|c||}{\textbf{Herz-Jesus-P8}}
  & \multicolumn{2}{|c|}{\textbf{Castle-P30}}\\ \cline{4-9}
  \multicolumn{3}{|l||}{~} & \textbf{Ours} & \textbf{VisualSFM}
  & \textbf{Ours} & \textbf{VisualSFM} & \textbf{Ours} & \textbf{VisualSFM}\\
  \hline
    \multirow{2}{5.5em}{Position error} & \multirow{2}{4.5em}{Meters} & mean & \textbf{0.0025} & 0.0027~ & \textbf{0.0042} & 0.0046~ & \textbf{0.0222} & 0.0316 \\ \cline{3-9}
  && median & \textbf{0.0026} & 0.0030~ & \textbf{0.0038} & 0.0043~ & \textbf{0.0210} & 0.0248 \\
  \hline
  \multirow{4}{5.5em}{Orientation error} & \multirow{2}{3em}{Frobenius} & mean & \textbf{~0.00065} & 0.00067 & \textbf{~0.00042} & 0.00054 & \textbf{~0.00098} & 0.0014 \\ \cline{3-9}
  && median & \textbf{~0.00056} & 0.00059 & \textbf{~0.00041} & 0.00051 & \textbf{~0.00099} & 0.0012 \\ \cline{2-9}
  & \multirow{2}{5.5em}{Degrees} & mean & \textbf{0.0263} & 0.0271~ & \textbf{0.0172} & 0.0219~ & \textbf{0.0399} & 0.0555 \\ \cline{3-9}
  && median & \textbf{0.0228} & 0.0241~ & \textbf{0.0166} & 0.0208~ & \textbf{0.0405} & 0.0478 \\
  \hline
\end{tabular}
}
\end{center}

\label{tab:pipeline}   
\vspace{-6mm}
\end{table*}
\subsection{Landmark tracking}
We simulated a landmark based robot positioning application. Our setup is described in Sec.~\ref{sec:introduction} (see Fig.~\ref{figure:robot_images}). Our goal is to recover the robot's position and orientation as it moves, where at each time step we only use the robot's current image and the two stationary images. To evaluate the methods we produced ``ground truth" measurements by taking 78 images of the scene from multiple locations using a single camera with fixed internal parameters. We calibrated the images using VisualSFM \cite{wu2011visualsfm}, utilizing their EXIF tags, and constraining  VisualSFM to share the same internal parameters across all the images. We further used the PTlens \cite{PTLens} software to remove  radial distortion from the images.  The  overall scale was adjusted to meters by triangulating points with known distance that could be seen from some of the images. The obtained internal calibration was used for all the cameras in the experiment. The exterior calibration parameters were used to determine the positions and orientations of the two stationary cameras in the experiment, and as ground truth measurements to evaluate our estimated positions and orientations of the robot for the 22 tested images.

Table \ref{tab:landmarks_33} shows the pose recovery errors obtained with our method both with 3 landmark points taken from each stationary camera and with 4 landmarks used from one stationary camera and 2 landmarks from the other camera. As the table shows, our method outperforms \cite{stewenius2005solutions,zheng2015structure,larsson2017efficient} and achieves comparable accuracies as \cite{byrod2009fast}, which is significantly slower than our method.

\subsection{RANSAC iterations}

\begin{figure}[tb]
\centering
\includegraphics[width=0.42\linewidth]{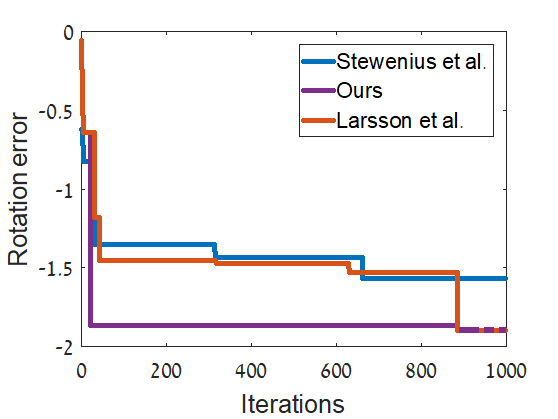} \ \ 
\caption{ \small
Camera orientation recovery using RANSAC. The Graph shows  rotation error  ($\mathrm{log}_{10}(\mathrm{degree}))$ as a function of the number of RANSAC samples.  
}
\vspace{-10pt}
\label{figure:ransac_graph}
\end{figure}

The next experiment demonstrates that our improved accuracy can affect the number of needed RANSAC iterations. For this experiment  we applied RANSAC to SIFT matches extracted from three images from the  Fountain-P11 dataset. We assume we know the parameters of two of the three cameras and use RANSAC with our method to compute the orientation of the third camera. Each RANSAC iteration selects three matching pairs from each of the known camera. Our method than produces 64 complex solutions from which we select the solution that minimizes the sum of Sampson errors for the six matching pairs. We finally plot for each RANSAC iteration the minimal error with respect to ground truth obtained up to that iteration. Fig.~\ref{figure:ransac_graph} shows the accuracy achieved with this procedure, as a function of RANSAC iteration, compared with the accuracies achieved when our method is replaced by \cite{stewenius2005solutions,larsson2017efficient}. It can be seen that our method achieved better accuracies in fewer iterations.

\subsection{Sequential structure from motion}

We finally demonstrates the utility of our method in a sequential multiview SFM pipeline. We produced a pipeline, similar to \cite{snavely2008modeling}, but with RANSAC implemented with our method. We begin with two images and use \cite{kukelova2008polynomial}'s implementation of the 5-points algorithm to compute the essential matrix between them. We then apply bundle adjustment \cite{triggs1999bundle} (using the SBA package~\cite{lourakis2009sba}) to obtain the two corresponding camera matrices. Next, for each additional image we use our method, with RANSAC, utilizing the previously computed cameras to robustly recover the exterior parameters of the new camera. This was followed by bundle adjustment, as in \cite{snavely2008modeling}, first using the already triangulated 3D points (using \cite{itseez2015opencv}), then triangulating new inlier matches, and, finally, removing points with large reprojection error.

We tested our pipeline on the  Herz-Jesus-P8, fountain-P11, and castle-P30 datasets \cite{strecha2008benchmarking}. We compare our pipeline  to  VisualSFM \cite{wu2011visualsfm}. For fair comparison we let both our method and VisualSFM use the same set of candidate matches (computed using \cite{wu2011siftgpu}) and the same order of images in both pipelines. Results are shown in Table~\ref{tab:pipeline}. Despite the use of matching pairs, instead of triplets, our method outperformed VisualSFM on all three datasets.

\vspace{-5pt}
\section{Conclusion}

We have introduced a novel solution to the 6 point, online problem of camera pose estimation from pairwise matches by constructing the Dixon resultant, which we implemented efficiently through symbolic derivations. We further provided a rigorous proof showing that there are 64 roots for the underlying polynomial system. We showed in experiments that our method obtains more accurate solutions than existing methods and showed the utility of the method in sequential SFM pipeline. We plan in future work to develop an analogous method for uncalibrated images.

\vspace{5pt}
{\small \noindent\textbf{Acknowledgment} Supported by the Minerva foundation with funding from the Federal German Ministry for Education and Research.} 

\clearpage

{\small
\bibliographystyle{ieee}
\bibliography{egbib}

\begin{thebibliography}{10}\itemsep=-1pt

\bibitem{PTLens}
{PTLens} automatic distortion correction.
\newblock \url{http://epaperpress.com/ptlens/}.

\bibitem{agarwal2011building}
S.~Agarwal, Y.~Furukawa, N.~Snavely, I.~Simon, B.~Curless, S.~M. Seitz, and
  R.~Szeliski.
\newblock Building rome in a day.
\newblock {\em Communications of the ACM}, 54(10):105--112, 2011.

\bibitem{ameller2000camera}
M.-A. Ameller, B.~Triggs, and L.~Quan.
\newblock Camera pose revisited--new linear algorithms.
\newblock In {\em European Conf. on Computer Vision (ECCV)}, 2000.

\bibitem{arie2012global}
M.~Arie-Nachimson, S.~Z. Kovalsky, I.~Kemelmacher-Shlizerman, A.~Singer, and
  R.~Basri.
\newblock Global motion estimation from point matches.
\newblock In {\em 3D Imaging, Modeling, Processing, Visualization and
  Transmission (3DIMPVT)}, pages 81--88, 2012.

\bibitem{bernshtein1975number}
D.~N. Bernshtein.
\newblock The number of roots of a system of equations.
\newblock {\em Functional Analysis and its Applications}, 9(3):183--185, 1975.

\bibitem{byrod2009fast}
M.~Byr{\"o}d, K.~Josephson, and K.~{\AA}str{\"o}m.
\newblock Fast and stable polynomial equation solving and its application to
  computer vision.
\newblock {\em Int. Journal of Computer Vision}, 84(3):237--256, 2009.

\bibitem{camposeco2017toroidal}
F.~Camposeco, T.~Sattler, A.~Cohen, A.~Geiger, and M.~Pollefeys.
\newblock Toroidal constraints for two-point localization under high outlier
  ratios.
\newblock In {\em Proc. CVPR}, volume~1, 2017.

\bibitem{camposeco2016minimal}
F.~Camposeco, T.~Sattler, and M.~Pollefeys.
\newblock Minimal solvers for generalized pose and scale estimation from two
  rays and one point.
\newblock In {\em European Conf. on Computer Vision}, pages 202--218. Springer,
  2016.

\bibitem{cox2006using}
D.~A. Cox, J.~Little, and D.~O'shea.
\newblock {\em Using algebraic geometry}, volume 185.
\newblock Springer, 2006.

\bibitem{dixon1909eliminant}
A.~L. Dixon.
\newblock The eliminant of three quantics in two independent variables.
\newblock {\em Proc. of the London Mathematical Society}, 2(1):49--69, 1909.

\bibitem{emiris1994sparse}
I.~Z. Emiris.
\newblock {\em Sparse elimination and applications in kinematics}.
\newblock PhD thesis, University of California, Berkeley, 1994.

\bibitem{gao2003complete}
X.-S. Gao, X.-R. Hou, J.~Tang, and H.-F. Cheng.
\newblock Complete solution classification for the perspective-three-point
  problem.
\newblock {\em IEEE Trans. on Pattern Analysis and Machine Intelligence},
  25(8):930--943, 2003.

\bibitem{hamilton1844ii}
W.~R. Hamilton.
\newblock On quaternions; or on a new system of imaginaries in algebra.
\newblock {\em The London, Edinburgh, and Dublin Philosophical Magazine and
  Journal of Science}, 25(163):10--13, 1844.

\bibitem{horn1991relative}
B.~K. Horn.
\newblock Relative orientation revisited.
\newblock {\em JOSA A}, 8(10):1630--1638, 1991.

\bibitem{itseez2015opencv}
Itseez.
\newblock Open source computer vision library.
\newblock \url{https://github.com/itseez/opencv}, 2015.

\bibitem{jiang2013global}
N.~Jiang, Z.~Cui, and P.~Tan.
\newblock A global linear method for camera pose registration.
\newblock In {\em Proc. of the IEEE Int. Conf. on Computer Vision (ICCV)},
  pages 481--488, 2013.

\bibitem{josephson2007image}
K.~Josephson, M.~Byrod, F.~Kahl, and K.~Astrom.
\newblock Image-based localization using hybrid feature correspondences.
\newblock In {\em Computer Vision and Pattern Recognition (CVPR)}, 2007.

\bibitem{kapur1994algebraic}
D.~Kapur, T.~Saxena, and L.~Yang.
\newblock Algebraic and geometric reasoning using dixon resultants.
\newblock In {\em Proc. of the Int. Symposium on Symbolic and Algebraic
  Computation}, pages 99--107, 1994.

\bibitem{kukelova2008automatic}
Z.~Kukelova, M.~Bujnak, and T.~Pajdla.
\newblock Automatic generator of minimal problem solvers.
\newblock In {\em European Conf. on Computer Vision}, pages 302--315, 2008.

\bibitem{kukelova2008polynomial}
Z.~Kukelova, M.~Bujnak, and T.~Pajdla.
\newblock Polynomial eigenvalue solutions to the 5-pt and 6-pt relative pose
  problems.
\newblock In {\em British Machine Vision Conf.}, volume~2, page 2008, 2008.

\bibitem{larsson2017efficient}
V.~Larsson, K.~Astr{\"o}m, and M.~Oskarsson.
\newblock Efficient solvers for minimal problems by syzygy-based reduction.
\newblock In {\em Computer Vision and Pattern Recognition (CVPR)}, 2017.

\bibitem{larsson2018beyond}
V.~Larsson, M.~Oskarsson, K.~{\AA}str{\"o}m, A.~Wallis, Z.~Kukelova, and
  T.~Pajdla.
\newblock Beyond gr{\"o}bner bases: Basis selection for minimal solvers.
\newblock {\em arXiv:1803.04360}, 2018.

\bibitem{lepetit2009epnp}
V.~Lepetit, F.~Moreno-Noguer, and P.~Fua.
\newblock {EPnP}: An accurate {O(n)} solution to the {PnP} problem.
\newblock {\em Int. Journal of Computer Vision}, 81(2):155--166, 2009.

\bibitem{li2006five}
H.~Li and R.~Hartley.
\newblock Five-point motion estimation made easy.
\newblock In {\em Int. Conf. on Pattern Recognition (ICPR)}, volume~1, pages
  630--633, 2006.

\bibitem{lourakis2009sba}
M.~I. Lourakis and A.~A. Argyros.
\newblock {SBA}: {A} software package for generic sparse bundle adjustment.
\newblock {\em ACM Trans. on Mathematical Software (TOMS)}, 36(1):2, 2009.

\bibitem{manocha1994solving}
D.~Manocha.
\newblock Solving systems of polynomial equations.
\newblock {\em IEEE Computer Graphics and Applications}, 14(2):46--55, 1994.

\bibitem{minimair15}
M.~Minimair.
\newblock {DR}: {Dixon} {Resultant} {Package} for {Maple}.
\newblock http://minimair.org/dr, 2015.

\bibitem{nister2004efficient}
D.~Nist{\'e}r.
\newblock An efficient solution to the five-point relative pose problem.
\newblock {\em IEEE Trans. on Pattern Analysis and Machine Intelligence},
  26(6):756--770, 2004.

\bibitem{snavely2008modeling}
N.~Snavely, S.~M. Seitz, and R.~Szeliski.
\newblock Modeling the world from internet photo collections.
\newblock {\em Int. Journal of Computer Vision}, 80(2):189--210, 2008.

\bibitem{stewenius2005solutions}
H.~Stewenius, D.~Nist{\'e}r, M.~Oskarsson, and K.~{\AA}str{\"o}m.
\newblock Solutions to minimal generalized relative pose problems.
\newblock In {\em OMNIVIS 2005}, 2005.

\bibitem{strecha2008benchmarking}
C.~Strecha, W.~Von~Hansen, L.~Van~Gool, P.~Fua, and U.~Thoennessen.
\newblock On benchmarking camera calibration and multi-view stereo for high
  resolution imagery.
\newblock In {\em Computer Vision and Pattern Recognition (CVPR)}, pages 1--8,
  2008.

\bibitem{sweeney2014gdls}
C.~Sweeney, V.~Fragoso, T.~H{\"o}llerer, and M.~Turk.
\newblock gdls: A scalable solution to the generalized pose and scale problem.
\newblock In {\em European Conf. on Computer Vision}, pages 16--31. Springer,
  2014.

\bibitem{triggs1999bundle}
B.~Triggs, P.~F. McLauchlan, R.~I. Hartley, and A.~W. Fitzgibbon.
\newblock Bundle adjustment: a modern synthesis.
\newblock In {\em Int. Workshop on Vision Algorithms}, pages 298--372, 1999.

\bibitem{verschelde1999algorithm}
J.~Verschelde.
\newblock Algorithm 795: Phcpack: A general-purpose solver for polynomial
  systems by homotopy continuation.
\newblock {\em ACM Trans. on Mathematical Software (TOMS)}, 25(2):251--276,
  1999.

\bibitem{wu2011siftgpu}
C.~Wu.
\newblock Siftgpu: A gpu implementation of scale invariant feature transform
  (sift)(2007).
\newblock {\em URL http://cs. unc. edu/\~{} ccwu/siftgpu}, 2011.

\bibitem{wu2011visualsfm}
C.~Wu.
\newblock Towards linear-time incremental structure from motion.
\newblock In {\em 3DTV}, pages 127--134, 2013.

\bibitem{zheng2015structure}
E.~Zheng and C.~Wu.
\newblock Structure from motion using structure-less resection.
\newblock In {\em Int. Conf. on Computer Vision (ICCV)}, pages 2075--2083,
  2015.

\end{thebibliography}
}

\end{document}